To appear in Acta Linguisica Academica, 2026

# Do Large Language Models Possess a Theory of Mind? A Comparative Evaluation Using the Strange Stories Paradigm

Anna Babarczy[1,2], András Lukács[3], Péter Vedres[2], Zétény Bujka[2]

1 Institute of General and Hungarian Linguistics, Research Centre for Linguistics

2 Department of Cognitive Science, Faculty of Natural Sciences, Budapest University of Technology and Economics

3 Institute of Mathematics, Eötvös Lóránd University

Abstract

The study explores whether current Large Language Models (LLMs) exhibit Theory of Mind (ToM) capabilities — specifically, the ability to infer others' beliefs, intentions, and emotions from text. Given that LLMs are trained on language data without social embodiment or access to other manifestations of mental representations, their apparent social-cognitive reasoning raises key questions about the nature of their understanding. Are they capable of robust mental state attribution indistinguishable from human ability in its output, or do their outputs merely reflect superficial pattern completion? To find an answer to this question, we tested five LLMs and compared their performance to that of human controls using an adapted version of a text-based tool widely used in human ToM research. The test consists in answering questions about the beliefs, intentions and emotions of story characters. The results revealed a performance gap between the models. Earlier and smaller models were strongly affected by the number of relevant inferential cues available and to some extent were also vulnerable to the presence of irrelevant or distracting information in the texts. In contrast, GPT-4o demonstrated high accuracy and strong robustness, performing comparably to humans even in the most challenging conditions. This work contributes to ongoing debates about the cognitive status of LLMs and the boundary between genuine understanding and statistical approximation.

## 1 Introduction

Large language models (LLMs) can produce responses that appear to reflect sophisticated social understanding, including the ability to reason about beliefs, intentions, emotions, and nonliteral meanings. This has prompted growing interest in whether their performance reflects a capacity resembling human Theory of Mind and pragmatic competence. However, high performance on established psychological tasks does not necessarily demonstrate robust social reasoning: models may exploit linguistic regularities, memorized task formats, or superficial cues that happen to correlate with the expected answers.

The present study examines the robustness of LLM performance on the Strange Stories task, a widely used assessment of advanced Theory of Mind and pragmatic inference. Rather than evaluating models only on the original stories, we systematically manipulated the information available in them. These manipulations allowed us to investigate how strongly model responses depended on the availability and salience of particular linguistic cues. We examined whether these effects differed across models of different types and sizes and how they compared to changes in human performance under the same experimental conditions.

## 2 Background

### *2.1 Large Language Models and Cognition*

Computational models for natural language processing (NLP) have been developed and evaluated since at least the 1950s, with successive paradigm shifts spanning symbolic approaches, statistical methods, and neural architectures. A major contemporary turning point was the introduction of the Transformer architecture (Vaswani et al., 2017), which has substantially influenced how NLP systems are trained and scaled.

Many prominent model families are derived from Transformer components. For example, BERT (Bidirectional Encoder Representations from Transformers) is an encoder-only model optimized for bidirectional representation learning (Devlin et al., 2019), whereas GPT-style (Generative Pre-trained Transformer) systems are decoder-only, autoregressive models designed for next-token prediction and text generation (Radford et al., 2018; Brown et al., 2020). Meta's LLaMA (Large Language Model Meta AI) family is likewise Transformer-based.

Modern large language models (LLMs) exhibit a broad range of natural language capabilities, including dialogue generation, summarization, translation, and various forms of problem solving and reasoning. Among these, dialogue generation and reasoning are particularly relevant to the present paper because they are often treated as outwardly "social-cognitive" behaviours: LLMs can produce coherent multi-turn responses, track conversational context, and adapt output style to interactional cues. They are often able to solve complex tasks that require logical inference, mathematical reasoning, and elements of strategic planning. These observations have prompted debates about the extent and nature of LLMs' underlying cognitive capabilities. Built on transformer architectures and trained on vast textual corpora, these models have begun to exhibit behaviours that resemble complex human faculties, raising the question – the focus of this paper - of whether such capacities are emergent properties of large-scale statistical learning or merely artefacts of linguistic pattern completion (e.g., Zhao et al., 2023, Xi et al., 2025).

### *2.2 Emergent Properties? Evidence from Reasoning*

While LLMs are fundamentally statistical models trained on vast corpora of text data, their performance on some tasks suggests the emergence of capabilities that mirror aspects of human cognition. These "emergent" properties appear abruptly or improve nonlinearly as model scale and training conditions change. Many of the emergent abilities observed in LLMs are framed in cognitive terms because they apparently enable these models to perform tasks traditionally associated with human cognition.

Sartori and Orrù, (2023), for instance, found that LLMs perform at the level of neurotypical adult humans in a variety of cognitive tasks including tests of reasoning and problem-solving. In a related effort, Wang et al. (2024) propose CogLM, a benchmark inspired by Piagetian developmental theory as a tool to assess the cognitive abilities of LLMs. The authors' report that advanced LLMs, such as GPT-4, achieve scores comparable to those of 20-year-old humans on parts of their test battery. Further evidence comes from Sandbrink and Summerfield (2024), who explore LLMs' and deep networks' ability to adapt to changing goals, rules, or contexts – a skill referred to as cognitive flexibility in human agents. The authors argue that LLMs fine-tuned via methods such as meta-reinforcement training and goal-directed policy training can demonstrate behaviours resembling human-like task switching, context adaptation, and generalization.

Such results are often taken as evidence that sophisticated, human-comparable behavioural performance can arise from large-scale language modelling. Skeptics argue, however, that proficient though these systems may seem at completing many real-world tasks, they still lack a robust, grounded understanding of the world and the language they generate. For example, Bender et al. (2021) characterize such models as "stochastic parrots" that are capable of mimicking statistical language patterns, but lack genuine semantic comprehension. On this view, LLM performance will be likely to degrade in settings that require deeper contextual understanding or flexible generalization—such as forms of inductive reasoning in which rules change, rules must be validated, or the task context deviates substantially from training distributions (Chen et al., 2024). Related critiques contend that many apparent successes may be explained by surface heuristics or memorized regularities rather than agent-like inference (Shapira et al., 2024; Ullman, 2023).

Several fine-grained evaluations show that "cognitive" competence is uneven across domains and can break down when tasks demand deeper causal, temporal, or social understanding. Some evidence comes from work on event knowledge, for instance. Kauf et al. (2023) test multiple pretrained LLMs using minimal sentence pairs. The authors report that several LLMs reliably distinguish possible from impossible events, but show weaker and less stable preferences when the distinction is between merely unlikely versus likely events. In the authors' interpretation, this suggests limits in how probabilistic real-world expectations are represented and generalized. They also report that generalization can vary by the type of linguistic change, with stronger consistency across some syntactic alternations than across semantically equivalent paraphrases. Miliani et al. (2025) introduce ExpliCa, a controlled benchmark for explicit reasoning that contrasts causal and temporal relations using overt connectives. The benchmark includes crowdsourced human acceptability ratings as a reference point. Across multiple evaluation setups, even strong contemporary models show substantial remaining error and often conflate temporal order with causal structure, with performance varying systematically as a function of linguistic presentation (e.g., iconic vs. anti-iconic ordering). In a temporally structured domain, Yuan et al. (2023) discuss their ExpTime (an instruction-tuning dataset for explainable temporal reasoning) and TimeLLaMA models fine-tuned from Llama-2, and present evidence that although instruction tuning can improve performance, it still leaves notable challenges for complex temporal reasoning and forecasting-style settings. Benchmarks targeting social and affective cognition further illustrate this unevenness. Sap et al. (2019) offer SocialIQA, a large-scale benchmark with about 38,000 multiple-choice questions assessing social commonsense reasoning, specifically targeting the ability to infer motivations, emotional reactions, and plausible consequences in human interactions. The first results showed a notable gap between model and human performance, particularly for questions requiring subtle inference or understanding of interpersonal dynamics. Sabour et al.'s (2024) EmoBench is a diagnostic benchmark assessing the emotional intelligence of LLMs. The authors' results indicate that while some models show near-human performance in emotional application (i.e., selecting socially appropriate actions in emotionally charged situations), they demonstrate systematic weaknesses in emotional understanding (i.e., inferring emotions or causes of emotions).

### *2.3 Theory of Mind as an Experience-Based Inferential Ability*

Within this broader debate, Theory of Mind (ToM)—the capacity to represent and reason about others' mental states such as beliefs, intentions and emotions—has become a central test case. ToM is usually treated not as a single, unitary faculty but as a constellation of related abilities (e.g., tracking different perspectives and interpretations of events, representing beliefs that may diverge from one's own reality, and integrating contextual cues in social interpretation). These abilities support the recognition that different agents can hold different mental representations of the same objective situation, which is foundational for social interaction, including communication, cooperation, and competition.

Several major theoretical traditions have been used to explain how ToM ability emerges in the mind. Theory-Theory proposes that people develop and repeatedly revise an intuitive, experience-driven "folk psychological" theory that links unobservable mental states to observable behaviour (Gopnik & Wellman, 1992). On this account, attributions of another person's internal states enable predictions about their future actions, and judgments about their likely choices. By contrast, Simulation Theory argues that social understanding is achieved by using one's own cognitive and affective systems as a model for others—i.e., by internally simulating what another agent might perceive, feel, or infer (Gallese & Goldman, 1998). In this view, ToM relies less on explicit "theorizing" and more on constructing an internal surrogate of the other's perspective and reasoning process. More recent computational approaches formalize ToM as probabilistic inference, often using Bayesian frameworks in which an observer updates beliefs about latent mental variables (e.g., beliefs, desires, intentions, preferences) given evidence in the form of observed behaviour (Baker et al., 2011). In broad terms, Bayesian inference combines prior expectations with the likelihood of the observed evidence under competing hypotheses to yield a posterior belief (an updated estimate of what is most plausible). Within computational ToM, mental-state attribution is treated as "inverse" inference: inferring hidden psychological causes from overt actions, rather than predicting actions from known mental states.

Across these perspectives, a recurring theme is that ToM is linked—developmentally and functionally—to rich interaction with the physical and social world. This motivates the question in the focus of this paper of whether systems that lack embodied experience and explicit social grounding can nevertheless approximate ToM-like reasoning in behaviourally measurable ways resembling an emergent property.

### 2.3.1 ToM Assessment Tasks

Theory of Mind is assessed using a diverse set of tasks that target different components (e.g., belief attribution, intention understanding, affective inference) and span multiple levels of difficulty from basic first-order reasoning to more advanced, contextually embedded mentalizing (Baron-Cohen et al., 1999; Shamay-Tsoory et. al., 2010; Heider and Simmel, 1944; Abell et. al., 2000).

A widely used test of ToM ability is Baron-Cohen et al.'s (1985) Sally-Anne "false-belief" task, in which one agent places an object in an initial location, leaves, and the object is then moved in the agent's absence; participants are asked where the agent will look for the object. This paradigm primarily assesses whether the participant can represent another person's belief when it conflicts with the participant's own beliefs, and it is typically administered as an explicit question-answer format, although variations exist that reduce verbal demands or use indirect response measures (Perner & Wimmer, 1985). False-belief tasks can also be extended to second-order reasoning by requiring participants to represent one agent's belief about another agent's belief (e.g., "A thinks that B thinks …"). Another classic first-order false belief task is the Unexpected Contents (Smarties) paradigm (Perner et. al., 1987): participants are shown a familiar container, report what they think is inside, discover that the contents are different from what they thought, and then judge what another person will believe is inside. In both of these classic tasks, performance is evaluated on a binary, pass or fail scale.

More Advanced ToM measures often use visual scenes, animations or verbal vignettes depicting socially complex situations that require participants to integrate pragmatic context, communicative intentions, and interpersonal norms. One prominent example for advanced vignette-based tasks is the Strange Stories task, originally developed by Happé (1994) to assess advanced mentalizing via short narratives that require participants to comprehend nonliteral language, social intentions, and context-dependent communicative acts and to reason through socially complex or ambiguous situations. These skills are all core components of higher-order mentalization crucial for navigating real-world social interactions. Participants' comprehension is tested via questions requiring them to explain the words or actions of a character or the causes of an event. The task has been widely used in both developmental and clinical research, and has also been adapted to address ceiling effects and to improve control over general comprehension and inference demands. In particular, White et al. (2009) describe a modified Strange Stories battery for schoolchildren that includes mental state stories alongside multiple types of control materials (including human, animal, and nature stories involving physical rather than mental inferences, as well as unlinked sentences measuring working memory) intended to better separate mentalizing-specific difficulties from more general, narrative comprehension or inferential difficulties.

Recent psychometric work suggests that Strange Stories performance may not reflect a single unified ToM factor, but instead may draw on a cluster of advanced social-cognitive and pragmatic competencies, which is important to acknowledge when interpreting scores as "pure" ToM. For this reason, Strange Stories can be especially informative for evaluating realistic mentalizing in language-capable systems (Nawaz et al. 2023).

Researchers have begun to apply several ToM task formats to LLMs and report mixed profiles of strengths and limitations across task types and models (e.g., Elyoseph et al., 2023). The next section provides a brief overview of these experiments.

### 2.3.2 Theory of Mind in LLMs

Integrating ToM capacities into artificial agents such as LLMs is widely viewed as an important objective for both researchers and developers, given ToM's central role in human social cognition. If such systems could reliably attribute and predict users' beliefs, intentions, and affective states, they could support more natural, context-sensitive, and cooperative human–AI interaction. In principle, robust ToM-like competence could improve intent recognition and response appropriateness across many application settings, particularly those involving extended dialogue and socially nuanced decision support.

Recent work has evaluated LLMs with ToM-relevant instruments and has reported a mix of strong performance on some benchmarks alongside substantial fragility under variation, prompting debates about what such outcomes do—and do not—imply about underlying mental-state reasoning. Evidence comparing LLM and human performance further suggests that apparent success can depend on task framing, prompt wording, and how ToM is operationalized in a given paradigm.

A prominent example is Kosinski's (2024) large-scale evaluation of false-belief reasoning in LLMs. In that study, 11 LLMs were tested on 40 false-belief tasks, and each task included closely matched true-belief controls plus reversed versions. The results showed a sharp gradient across model generations: older models solved no tasks, GPT-3-davinci-003 and ChatGPT-3.5-turbo solved about 20% of tasks, and ChatGPT-4 solved 75%, which the paper reports as comparable to the performance of six-year-old children in prior human studies. Kosinski interprets these findings as consistent with the possibility that ToM-like behaviour could emerge as an unintended by-product of improving language competence, while also acknowledging alternative explanations (e.g., strategies not constituting "human-like" ToM).

Beyond false-belief tasks, complex vignette-based measures (including Strange Stories) have also been used to probe LLM mentalizing, often as part of larger batteries rather than as a focus of systematic experimental manipulation. For example, Strachan et al. (2024) compared humans with GPT-4, GPT-3.5, and LLaMA2-70B across a broad set of ToM tests and found that GPT-4 performed at or above human levels on many measures (including Strange Stories), while faux pas detection was a relative weakness for GPT models in their battery. In the same study, LLaMA2-70B generally performed worse than the GPT models but showed an apparent advantage on faux pas; follow-up analyses suggested this advantage could reflect task-specific biases (e.g., a tendency to attribute ignorance) rather than robust mentalizing. Related work has examined ToM performance in a more qualitative, case-study style. Brunet-Gouet et al. (2023) report that (versions of) ChatGPT can produce plausible answers on several language-based ToM tasks, including Strange Stories, while also showing sensitivity to prompting. Chen et al. (2024) also took inspiration from the Strange Stories test. They created 201 new stories modelled on the original ones and converted the task into a multiple-choice test. Several other ToM measures were also adapted. Their results showed that in the Strange Stories part of their battery, GPT4 performed at near-human level and other models' performance could also be improved through step-by-step (Chain-of-Thought) prompting.

A major counterpoint in this literature is that ToM-like performance can be brittle and may rely on shortcuts. Shapira et al. (2024) explicitly frame many LLM behaviours as "Neural Theory of Mind (N-ToM)" and report that performance often degrades under minor manipulations, consistent with reliance on shallow heuristics rather than stable belief modelling. The authors compare this behaviour to the phenomenon termed the Clever Hans effect, whereby a system (animal, human, or machine) appears to demonstrate intelligence or understanding, but is in fact responding to subtle, unintended cues in the environment rather than truly solving the task at hand. Ullman (2023), for instance, demonstrates that LLM performance on standard false-belief paradigms can be brittle, with apparent success hinging on superficial properties of task presentation rather than stable belief attribution. Using GPT-3–class systems, the study tests variants of the Sally–Anne false-belief scenario that introduce only minimal manipulations—such as making a container transparent, switching which agent's belief is queried, or making slight changes to prompt wording. These alterations, which should not systematically mislead typical human participants, nevertheless produce large and consistent drops in model accuracy. Ullman interprets this fragility as evidence that correct answers in canonical versions may often be driven by learned textual heuristics and prompt-contingent pattern completion, rather than robust reasoning about another agent's perspective.

Pi et al. (2024), however, offer an analysis of why LLMs sometimes fail at modified false-belief tasks. The authors introduce SCALPEL, a method for making inferences explicit through targeted prompt modifications. Their results show that some failures can be mitigated when prompts explicitly cue bridging inferences, suggesting that breakdowns may sometimes reflect gaps in commonsense inference rather than an absence of mental-state representation.

A similar conclusion was reached by van Duijn et. al. (2023), who used the original Strange Stories as their baseline condition and created two additional versions for comparison: (a) a superficially modified variant in which surface details (e.g., names, locations) were changed while the core structure was preserved, and (b) a novel scenario written to target the same underlying ToM construct in a different narrative context. They evaluated four base (non-instruction-tuned) LLMs and seven instruction-tuned LLMs, and compared model performance with that of 7–10-year-old children. Their results indicate that instruction-tuned models were largely stable across the two modified versions, whereas base models showed a small decline in performance under modification; overall, instruction-tuned models matched or exceeded child performance, while base models performed below the level observed for 7–8-year-olds.

In summary, across recent evaluations, LLMs can appear to perform well on established ToM tasks, yet this performance is often highly susceptible to minor changes in task wording, framing, or other superficial features. This fragility complicates interpretation, because a "pass" on a canonical item may not generalize to near-equivalent versions that leave the underlying mental-state inference requirements unchanged. Although some work shows that providing explicit guidance (e.g., instructing the model to reason step-by-step or to attend to an agent's perspective) can partially counteract such failures and improve accuracy, reliance on increasingly explicit instructions is not a satisfactory solution from an applied perspective. If robust mental-state attribution must be repeatedly prompted, this shifts the burden onto users and designers, reducing the efficiency and naturalness of human–LLM interaction in realistic settings. Just as importantly, instruction-dependent improvements do not resolve the theoretical question at the centre of this literature: whether ToM-like outputs reflect an emergent, generalizable capacity for mental-state reasoning, or whether they are better understood as sophisticated but brittle pattern mimicry that succeeds when tasks contain the right cues.

For these reasons, the present study moves beyond single-form benchmark administration and tests LLM mentalising under systematic task modifications. Building on the Strange Stories paradigm, the planned manipulations are designed not only to reduce superficial familiarity, but also to increase inferential complexity in controlled ways, thereby probing whether LLM performance remains stable when successful responding requires deeper, less cue-dependent mental-state attribution.

## 3 Method

### *3.1 LLMs and Human Participants*

Five LLMs of differing sizes, ChatGPT-3.5 turbo, ChatGPT-4o, Gemma 2, LLaMA 3.1 and Phi 3 were included in the study. All five models had a temperature setting of 0.5 for testing. The main specifications of the models are listed in Table 1. The language of testing was English. For human controls, 30 university students (7 female) participated in the experiment. All spoke Hungarian as their native language but were fluent speakers of English with advanced level certificates recognized in Hungary. The study received ethical approval from the Ethical Review Board of the Budapest University of Technology and Economics. Participants completed the study online and provided informed consent before beginning; no personally identifiable information was collected.

| LLM | Number of Parameters | Deployment | Developer |
|---|---|---|---|
| ChatGPT-3.5 turbo | undisclosed[1] | Managed API | OpenAI |
| ChatGPT-4o | undisclosed | Managed API | OpenAI |
| Gemma 2 | 9 billion | Local | Google |
| LLaMA 3.1 | 8 billion | Local | Meta AI |
| Phi 3 | 14 billion | Local | Microsoft Research |

Table 1. Main properties of the LLMs included in the study

*3.2 Materials and Procedures*

The updated (White et al., 2009) version of Happé's (1994) Strange Stories task was adapted for the purposes of the study. The White et al. versions of three categories of stories were used as baseline: 8 animal stories, 8 mental state stories and 12 physical state stories. In an effort to control for the effects of real-world knowledge, this set was supplemented by 6 new animal stories involving fictitious behaviours of fictitious animals. One open-ended ”why” or “how” question asking for the justification of a behaviour or event in the story accompanied each text. In the animal category, the test question referred to the behaviour of an animal in the story, while in the mental state category, the question concerned the mental state of one of the human characters. The physical state stories tested causal inferences not involving mental states. Each baseline story was about 80 to 120 words long. An example for each category of baseline story is shown in Table 2. The complete set of fictitious animal stories written for the purposes of this study is given in Appendix A.

| Category | Story with Question |
|---|---|
| Animal<br>(N = 8) | Animals that live in groups often have an order of importance within the group. The strongest male is the leader of the group. This leader will often attack other animals in the group who are not as strong as this leader. This shows the other animals how important the leader is. In a group of chickens a very small chicken hasn't got many of its feathers left. QUESTION: Why hasn't this chicken got many feathers left? |
| Fictitious Animal<br>(N = 6) | Many extremely steep mountains cover an island. Above a certain height no animal can stay conscious, because of the lack of oxygen. Floots are the most nimble and shifty species around. They are the only ones capable of climbing above a certain height, because of their special hooves. They will, however, often get drowsy during their climbs. Scavengers in the area often feast on floots, even though they are very poor climbers. They can do this without hunting them. QUESTION: How is that possible? |
| Mental State<br>(N = 8) | During the war, the Red army captures a member of the Blue army. They want him to tell them where his army's tanks are; they know they are either by the sea or in the mountains. They know that the prisoner will not want |

[1] Although the number of parameters has not been disclosed for ChatGPT-3.5 turbo, the predecessor model family on which the GPT-3.5 series builds, was reported to contain up to 175 billion parameters (Brown et al., 2020).

| | |
|---|---|
| | to tell them, he will want to save his army, and so he will certainly lie to them. The prisoner is very brave and very clever, he will not let them find his tanks. The tanks are really in the mountains. Now when the other side asks him where his tanks are, he says, "They are in the mountains." QUESTION: Why does he say that? |
| Physical State<br>(N = 12) | A burglar is about to break into a jewellers' shop. He skilfully picks the lock on the shop door. Carefully he steps over the electronic detector beam. If he breaks this beam, it will set off the alarm. Quietly he opens the door of the storeroom and sees the gems glittering. As he reaches out, however, he steps on something soft. He hears a screech and something small and furry runs out past him. Immediately the alarm sounds. QUESTION: Why did the alarm go off? |

Table 2. An example of the baseline version of each category of the Strange Stories task.

We modified the baseline stories in two directions as shown in Table 3: one involved the reduction of information constituting potential inferential cues. The original stories in White et al. (2009) are constructed in a way that information relevant for the social or physical inference to be made is reiterated. The key clues are mentioned multiple times to direct the reader's attention to them. As a first level of reduction, these reiterations were removed. As a second level of information reduction, other leading qualities of the stories were also removed, such as explicit mentions of mental states and function words signalling cause-and-effect relationships. The other direction of modification involved the addition of two pieces of irrelevant, potentially distracting information. One piece of irrelevant information concerned the human or animal agent or central inanimate object in the story, while the other piece stated some inconsequential fact about a prominent object or concept.

Both types of modification had three levels resulting in a total of seven versions of each story. In this paper, we shall omit the discussion of intermediate levels and focus on the final information-reduced version, and the final expanded (distraction) version. The questions accompanying the stories did not change across story versions.

| Modification Type | Modified Story |
|---|---|
| Reduced Information<br>(The phrases ~~crossed out~~ here were removed from the baseline story.) | During the war, the Red army captures a member of the Blue army. They want him to tell them where his army's tanks are; ~~they know they are~~ either by the sea or in the mountains. ~~They know that~~ The prisoner ~~will not want to tell them, he will want to save his army, and so he will certainly lie to them~~. ~~The prisoner is very brave and very clever, he~~ will not let them find his tanks. The tanks are really in the mountains. Now when the other side asks him where his tanks are, he says, "They are in the mountains." |
| Distracting Information<br>(The phrases appearing in **bold** here were | During the war, the Red army captures a member of the Blue army. **During the** |
| added to the baseline story.) | **fighting the prisoner suffered a serious concussion**. They want him to tell them where his army's tanks are; they know they are either by the sea or in the mountains. **The Blue army's tanks are very quick and can relocate very quickly.** They know that the prisoner will not want to tell them, he will want to save his army, and so he will certainly lie to them. The prisoner is very brave and very clever, he will not let them find his tanks. The tanks are really in the mountains. Now when the other side asks him where his tanks are, he says, "They are in the mountains." |

Table 3. Hand-modified versions of a mental state story from the Strange Stories task.

All open-weight language models were obtained and run locally using the Ollama software platform (version 0.6.0). Models were deployed in isolated Docker (version 27.3.1) containers to ensure that each model operated in a consistent Linux-based computational environment, with standardized hardware access and software dependencies across all sessions. Experiments were conducted on a personal computer equipped with an AMD Ryzen 9 4900HS processor (3.00 GHz, 8 cores) with integrated Radeon Graphics and 16 GB of RAM. Interaction with the models was automated in R (version 4.4.0; R Core Team, 2024) via the RStudio integrated development environment (version 2024.09.1 Posit, PBC, 2024). The models were tested on all seven versions of the stories, with five iterations of each trial. These five stochastic runs were treated analogously to participants as explained in the statistical model descriptions in the Results section. For each trial, models received the following instruction: "I will provide you with a short story. Based on the story, answer the question provided at the end. Include your reasoning." This instruction was followed by one of the stories, then came the word "QUESTION:" followed by the question.

Human participants were randomly assigned to one of three groups: One group completed the task in the Baseline condition, one in the Reduced Information condition, and the third group read the stories in the Distracting Information condition. Human controls completed the test implemented in Python and running online on Pavlovia.org. They saw one story at a time on the screen. The order of the stories was automatically randomised to counterbalance order and fatigue effects. The question was printed below the story. When participants were ready to answer, they pressed the *continue* button, which led them to the next screen with the question printed on top and a text box appearing in the lower half. Participants typed their answer into the box, and moved on to the next trial by pressing the *submit* button.

All answers were hand-scored by a human rater on a scale of 0 to 2 in accordance with the instructions provided in the Appendix of White et al. (2009) for their Strange Stories task. Fully correct answers were given 2 points, partially correct answers were given 1 point and incorrect answers were given 0 points. Responses were classified as incorrect if they were unrelated to the story's context, demonstrated a flaw in reasoning, or included factual inaccuracies based on the information provided in the text. For the mental state story we are using here as an example the scoring guidelines specify that

- 2 points—reference to fact that other army will not believe and hence look in other place, reference to prisoner's realization that that's what they'll do, or reference to double bluff
- 1 point—reference to outcome (to save his army's tanks) or to mislead them
- 0 points—reference to motivation that misses the point of double bluff (he was scared)

Analogous scoring guidelines were created for the fictitious animal stories written for the purposes of our study. For the story shown in Table 2, the instructions are

- 2 points—reference to fact that floots get drowsy if they climb high and since the terrain is steep, they may fall. Scavengers may exploit this
- 1 point—reference to drowsiness or the terrain without linking them to make inferences
- 0 points—any other answer, such as scavengers learning to climb

**4 Results**

Inter-rater agreement was assessed on a balanced subset of the data. To ensure comparable representation across responder types, modification conditions, and story types, we used stratified random sampling with 11 responses selected from each combination of human/LLM group, modification condition, and story type. This yielded a balanced inter-rater reliability sample of 1,080 responses corresponding to 30.25% of the 3,570 responses in the data set.

The 1,080 responses were scored independently by two human raters. Both raters were blind to the modification condition and to the identity and group of the responder. Scoring followed the guidelines provided in the Appendix in White et al. (2009) for the original Strange Stories items. For the fictitious animal stories created for the purpose of this study, analogous scoring guidelines were prepared by a researcher who was not involved in the rating process.

Inter-rater agreement was quantified using weighted Cohen's kappa suitable for our ordinal scoring system. Linear weighted kappa was treated as the primary index of agreement, with exact agreement and quadratic weighted kappa also reported for completeness. Of the 1080 pairs of ratings, the two raters gave the exact same score on 989 responses, corresponding to 91.57% exact agreement. Cohen's kappa weighted to reflect the distance between non-matching scores was high with linear weighted $\varkappa = .866$, and quadratic weighted $\varkappa = .904$ indicating very strong inter-rater agreement.

Statistical analyses relating to our research questions were carried out using R (version 4.5.2; R Core Team, 2025) via the RStudio integrated development environment (version 2025.09.2; Posit team, 2025) Plots were created using ggplot. For statistical analyses, the five iterations (stochastic runs) of trials for LLMs were treated as the source of unsystematic variability within that LLM. The statistical models were fitted with Score (0, 1 or 2 points) or mean Score as the dependent variable, LLM (GPT-3.5, GPT-4o, Gemma 2, LlaMA 3.1, Phi 3 and Human) as a between-subjects independent variable, Story Type (Animal, Fictitious Animal, Mental and Physical) as a within-subjects independent variable, and Modification Type (Baseline, Reduced, Distracting) as a within-subjects independent variable for the AIs and a between-subjects independent variable for humans.

*4.1 The Effects of Story Type*

Our first analysis examines differences between LLMs and the effects of Story Type on mean scores across Modification Type. The mean scores for each Story Type and LLM/Human across all modification types are shown in Figure 1.

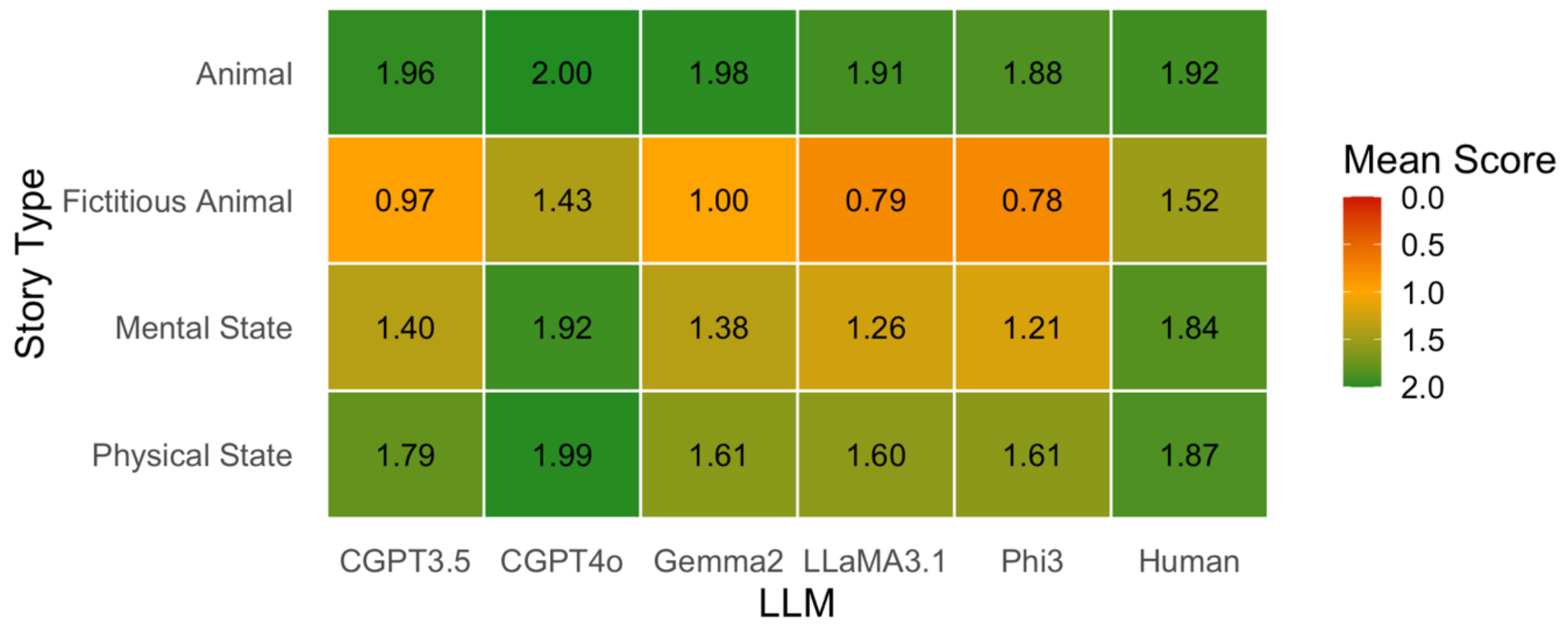

Figure 1: Mean scores by Story Type and LLM across all modification types.

A 6 (Model) x 4 (Story Type) ANOVA revealed a significant main effect of LLM ($F(5, 49) = 58.71$, $p < .001$), a significant main effect of Story Type ($F(3, 147) = 377.53$, $p < .001$) and a significant interaction between LLM and Story Type ($F(15, 147) = 17.70$, $p < .001$). Planned contrasts comparing individual AIs' performance to humans' across story types revealed that ChatGPT 4o did not differ from human controls ($t(49) = 1.34$, $p = .186$) but all other LLMs showed significantly weaker inferential competence ($p < .001$ for all). Effect sizes were substantial ranging from Cohen's $d = 2.10$ for ChatGPT-3.5 to $d = 3.40$ for Phi 3. Another set of planned contrasts using trend analysis established the order of difficulty of Story Types across all LLMs. Animal Stories proved to be the easiest to interpret, they were followed by the Physical State stories ($d = 1.61$), Mental State stories came next ($d = 1.97$) and finally, with an extensive gap, Fictitious Animal stories ($d =3.41$) last (see Appendix B for some incorrect answers in this category). We attribute the almost flawless performance for animal stories to the circumstance that responders could rely on real-world knowledge t a great extent.

As indicated by the mean scores in Figure 1, the significant interaction is explained by the finding that while for human controls and ChatGPT-4o, only the Fictitious Animal stories reduced accuracy to a substantial degree (M = 1.52 and 1.43 respectively) with the remaining three Story Types showing similar levels of difficulty (Means of 1.84 and above), for the older and smaller AIs there was a clear decline in performance between every two consecutive Story Types in the ranking.

*4.2 The Effects of Modification*

Our second analysis concerns the effects of information reduction and of the introduction of distracting information on inferential ability. Mean scores by LLM and Modification Type across all story types are shown in Figure 2.

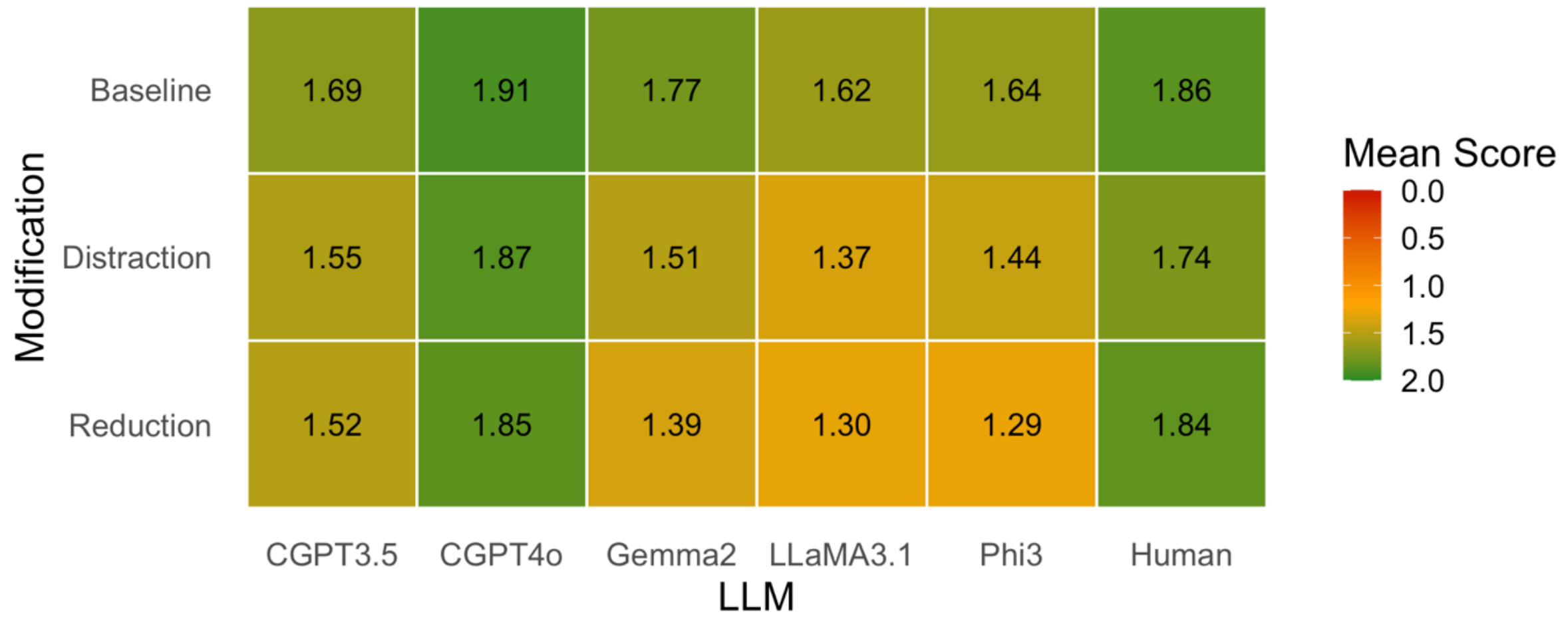

Figure 2: Mean scores by LLM and Modification Type across all story types

A series of ordinal (cumulative link) mixed-effects models were fitted with Laplace approximation to investigate the effects of LLM (between-subjects), Modification Type (repeated-measures for AIs and between-subjects for humans), and their interaction on response scores. The base model included random intercepts for individual story items, and for LLM stochastic runs/human participants, where each stochastic run or participant was identified by a unique iteration ID (AIs) or by a participant ID (humans). The fixed factors of LLM and Modification Type were added next, and finally their interaction. Both fixed factors were sum coded. Improvement in model fit with the addition of fixed factors was tested using the Likelihood Ratio test.

Compared to the intercept-only model, the addition of LLM significantly improved model fit ($\chi^2(5) = 100.54$, $p < .001$), as did the addition of Modification Type ($\chi^2(2) = 124.86$, $p < .001$), and the addition of the interaction term ($\chi^2(10) = 43.12$, $p < .001$). The final model results displayed in Table 4 reveal that all AIs except ChatGPT-4o performed significantly worse than average (p = or < .001), and ChatGPT-4o performed significantly better than average (p < .001). On the whole, both information reduction and the introduction of distracting information negatively affected the LLMs' ability to draw inferences from the stories (p < .001 for both). Overall, the reduction of inferential cues had a much stronger effect (b = -0.46) than the addition of irrelevant information (b = -0.27). The interaction terms further reveal that Gemma 2 and Phi 3 were especially vulnerable to the reduction of information in the stories (b = -0.42 and -0.34 respectively). The remaining LLMs' susceptibility did not differ significantly from average although the fairly large positive interaction coefficients for ChatGPT-4o suggest that it may be more resilient than other LLMs.

| **Term** | **Estimate (b)** | **SE** | **z value** | **p** |
|---|---|---|---|---|
| **LLM** | | | | |
| CGPT-3.5 | -0.34 | 0.11 | -3.23 | .001 |
| CGPT-4o | 1.68 | 0.14 | 11.87 | < .001 |
| Gemma 2 | -0.42 | 0.11 | -3.93 | < .001 |
| LLaMA 3.1 | -1.06 | 0.10 | -10.21 | < .001 |
| Phi 3 | -0.91 | 0.10 | -8.65 | < .001 |
| Human | 1.04 | | | |
| **Modification** | | | | |
| Distraction | -0.27 | 0.07 | -3.91 | < .001 |
| Reduction | -0.46 | 0.07 | -6.82 | < .001 |
| Baseline | 0.73 | | | |
| **LLM by Modification Interaction** | | | | |
| CGPT-3.5:Distraction | 0.10 | 0.15 | 0.71 | .477 |
| CGPT-4o:Distraction | 0.20 | 0.19 | 1.05 | .295 |
| Gemma 2:Distraction | -0.05 | 0.15 | -0.31 | .756 |
| LLaMA 3.1:Distraction | -0.03 | 0.14 | -0.21 | .835 |
| Phi 3:Distraction | 0.16 | 0.14 | 1.14 | .253 |
| Human:Distraction | -0.40 | | | |
| CGPT-3.5:Reduction | 0.07 | 0.15 | 0.48 | .634 |
| CGPT-4o:Reduction | 0.18 | 0.19 | 0.97 | .332 |
| Gemma 2:Reduction | -0.42 | 0.15 | -2.91 | .003 |
| LLaMA 3.1:Reduction | -0.16 | 0.14 | -1.14 | .256 |
| Phi 3:Reduction | -0.34 | 0.14 | -2.42 | .015 |
| Human:Reduction | 0.67 | | | |

Table 4. Model coefficients, Standard Errors, z values and significance values. clmm(score ~ LLM * Modification + (1|participant) + (1|item), data = dat, link = "logit", Hess = TRUE)

To further investigate the significant interaction, each LLM's susceptibility to Modification was tested using estimated marginal means from the cumulative link model. For each model, planned contrasts tested performance differences between Distraction vs Baseline and between Reduction vs Baseline (Wald z tests, df = Inf), with p-values adjusted using Holm correction. The results are shown in Table 5. As we can see, all AIs except ChatGPT-4o, as well as humans performed significantly worse in the distracting information condition than with the Baseline stories. Reasoning ability also declined significantly for all but ChatGPT-4o and human controls when the amount of information was reduced in the stories.

| **LLM** | **Estimate** | **SE** | **z ratio** | **p** | **Estimate** | **SE** | **z ratio** | **p** |
|---|---|---|---|---|---|---|---|---|
| | **Contrast: Distraction vs Baseline** | | | | **Contrast: Reduction vs Baseline** | | | |
| CGPT3.5 | -0.72 | 0.29 | -2.51 | .012 | -0.95 | 0.28 | -3.35 | .002 |
| CGPT4o | -0.41 | 0.40 | -1.02 | .308 | -0.62 | 0.39 | -1.61 | .217 |
| Gemma2 | -1.51 | 0.30 | -5.00 | <.001 | -2.08 | 0.30 | -6.99 | <.001 |
| LLaMA3.1 | -1.21 | 0.27 | -4.41 | <.001 | -1.53 | 0.27 | -5.64 | <.001 |
| Phi3 | -1.00 | 0.29 | -3.53 | .004 | -1.70 | 0.28 | -6.15 | <.001 |
| Human | -1.12 | 0.24 | -4.64 | <.001 | -0.25 | 0.26 | -0.99 | .324 |

Table 5. LLM susceptibility to modifications computed as marginal means differences (Estimate) between Modification Types.

Finally, we used the differences in estimated marginal means shown in Table 5 to compare each AI's modification susceptibility to human susceptibility. The planned contrasts used Wald z tests (df = Inf) on the model's link (logit) scale; p values were adjusted using Holm correction. The results of the analysis are shown in Table 6.

| **LLM Pair** | **Estimate** | **SE** | **z ratio** | **p** | **Estimate** | **SE** | **z ratio** | **p** |
|---|---|---|---|---|---|---|---|---|
| | **Distraction vs Baseline** | | | | **Reduction vs Baseline** | | | |
| CGPT3.5 - Human | 0.40 | 0.37 | 1.08 | 1.000 | -0.70 | 0.38 | -1.82 | .136 |
| CGPT4o - Human | 0.71 | 0.47 | 1.53 | .632 | -0.37 | 0.46 | -0.80 | .425 |
| Gemma2 - Human | -0.39 | 0.39 | -1.01 | 1.000 | -1.83 | 0.39 | -4.66 | <.001 |
| LLaMA3.1 - Human | -0.09 | 0.36 | -0.24 | 1.000 | -1.28 | 0.37 | -3.43 | .002 |
| Phi3 - Human | 0.12 | 0.37 | 0.32 | 1.000 | -1.45 | 0.38 | -3.85 | .001 |

Table 6. Comparison of AI susceptibility and Human susceptibility to the introduction of distracting information and to the reduction of inferential cues. Estimate: Difference between differences.

The analysis explicates the results of the cumulative link mixed model. None of the AI – Human comparisons involving vulnerability to distracting information is significant. That is, it is now clear that although distracting information somewhat reduces performance, it does not interfere with LLM reasoning any more than it does with human reasoning. The reduction of relevant information, in contrast, does affect the reasoning abilities of Gemma 2 (Estimate = -1.83), LLaMA 3.1 (Estimate = -1.28) and Phi 3 (Estimate = -1.45) to a significantly greater extent than those of humans; all of these comparisons are significant ($p < .001$, $p = .002$ and $p = .001$ respectively). Both GPT models proved just as resilient as humans, however.

**5 Summary and Conclusion**

The results of the analysis testing differences between story types regardless of modification condition confirmed our expectations and reinforced previous findings: Encyclopaedic knowledge aids the inferential process for both humans and LLMs, while the intricacies of social conventions are more difficult for most LLMs to navigate. Chat GPT 4o, however, performed at or near ceiling and matched human performance even on the more challenging tasks of reasoning about fictitious animals and drawing inferences about human states of mind. These findings are consistent with previous research suggesting that large language models can achieve human-like accuracy on standard ToM assessments. However, performance on standard tasks offers little insight into whether these results stem from genuinely sophisticated reasoning abilities or from refined heuristics and mimicry.

The story modifications introduced in this study provided a means of testing the flexibility of LLM reasoning. If language models are indeed on par with humans in their ToM abilities, then minor alterations to task structure and information content should not meaningfully affect their performance. One type of modification was the reduction of redundance and semantic information, which limited the amount of contextual cues available for reasoning. GPT-4o, the strongest-performing model, showed impressive resilience to this manipulation. Although it exhibited small declines in performance across some story types, none of these drops reached statistical significance. This robustness supports the interpretation that GPT-4o engages in stable, context-sensitive inferential reasoning and lends preliminary support to the claim that ToM-like abilities may be present in the model. By contrast, ChatGPT-3.5, Gemma 2, LLaMA 3.1, and Phi 3 showed significantly greater vulnerability to increased inferential complexity when compared to human controls. These models performed well when semantic cues were rich and relevant but deteriorated sharply under abstraction. This pattern suggests a superficial form of ToM, where surface-level proficiency on ToM tasks may mask a reliance on heuristic shortcuts. When deprived of guiding semantic markers, these models failed to sustain their performance, indicating that their apparent mentalization ability may not generalize beyond familiar conditions.

A second type of manipulation involved the introduction of semantic distractors designed to test the models' resistance to irrelevant but potentially misleading contextual information. Both humans and AIs — with the exception of GPT-4o — showed a significant decline in performance under this condition compared to the baseline. The effect was no more pronounced for AIs than it was for humans, however. This finding suggests that the larger models were generally able to maintain focus on relevant inferential pathways despite increased semantic noise but the three smallest models, similarly to humans, were less resilient.

Taken together, these results indicate that apparent ToM competence in LLMs is best characterized in terms of robustness under perturbation rather than accuracy on canonical items alone. The fact that multiple models performed well when semantic cues were rich, yet deteriorated under reduced redundancy/abstraction and under semantic distractors, suggests that at least part of their success on standard ToM-style vignettes can be supported by cue-dependent strategies that do not reliably generalize. In contrast, GPT-4o's comparatively stable performance across both manipulations is consistent with the possibility that some contemporary models can sustain more context-sensitive inference across moderately altered conditions, although this study does not on its own establish that the underlying mechanism is the same as human mental-state reasoning.

When interpreting the results, we should also keep in mind that for humans, susceptibility to cue degradation or distraction was inferred from differences between independent groups of participants. For LLMs, susceptibility was inferred from performance differences within the same model across versions of the same story items. Therefore, the comparison between humans and LLMs is not a perfectly matched within-subject comparison and it should be interpreted with this design difference in mind.

A cautious interpretation is therefore that language-only training can, in some cases, provide sufficiently rich statistical and relational structure for powerful learning systems to build internal, usable approximations of world knowledge and social regularities, enabling outputs that resemble human-like mental-state attribution in many vignette-based settings. Natural language encodes a dense record of human goals, conflicts, norms, and explanations, and large-scale exposure may allow models to extract patterns that support surprisingly sophisticated inferences. These findings can be situated within the broader interpretive debate exemplified by Kosinski's (2023) influential evaluation of LLMs on a large battery of Theory of Mind–relevant tasks. Kosinski adopts a relatively optimistic interpretation, arguing that ToM-like task performance may plausibly arise as an emergent by-product of large-scale language modelling rather than from explicit architectural mechanisms designed for mental-state reasoning. On this functionalist reading, consistently ToM-indistinguishable behavioural outputs provide grounds to take the possibility of ToM-like capacity seriously, even if the underlying causal mechanisms differ from those in humans. At the same time, the present results highlight why this functional criterion must be supplemented with robustness testing. Accordingly, these findings motivate the methodological conclusion that evaluating LLM mentalising requires advanced ToM tasks with systematic, theory-driven modifications that go beyond surface familiarity reductions and directly manipulate inferential demands. In this study, the combination of cue-reduction (increased abstraction) and distractor-based perturbations already differentiates models that appear strong on standard items from those that maintain performance when the inferential pathway is less explicitly scaffolded by the text. Extending this approach with Strange Stories variants that more deliberately increase inferential complexity provides a clearer test of whether LLM mental-state attribution is robust and generalizable, or primarily dependent on favourable linguistic packaging.

An important direction for future research is to extend the present comparisons to recent reasoning-specialised models. Such models may differ from standard large language models not only in overall performance, but also in the strategies they use when pragmatic interpretation requires sensitivity to speaker intentions, multi-step inference from degraded cues, or the integration of conflicting contextual cues.

References

Abell, F., Happé, F., & Frith, U. (2000). Do triangles play tricks? Attribution of mental states to animated shapes in normal and abnormal development. *Cognitive Development, 15*(1), 1–16. https://doi.org/10.1016/S0885-2014(00)00014-9

Baker, C. L., Saxe, R., & Tenenbaum, J. B. (2011). Bayesian theory of mind: Modeling joint belief–desire attribution. *Proceedings of the Annual Meeting of the Cognitive Science Society, 33*. http://web.mit.edu/9.s915/www/classes/theoryOfMind.pdf

Baron-Cohen, S., Leslie, A. M., & Frith, U. (1985). Does the autistic child have a "theory of mind"? *Cognition, 21*(1), 37–46. https://doi.org/10.1016/0010-0277(85)90022-8

Baron-Cohen, S., O'Riordan, M., Stone, V. E., Jones, R., & Plaisted, K. (1999). A new test of social sensitivity: Detection of faux pas in normal children and children with Asperger syndrome. *Journal of Autism and Developmental Disorders, 29*(5), 407–418. https://doi.org/10.1023/A:1023035012436

Bender, E. M., Gebru, T., McMillan-Major, A., & Mitchell, M. (2021). On the dangers of stochastic parrots: Can language models be too big? In *Proceedings of the 2021 ACM Conference on Fairness, Accountability, and Transparency* (pp. 610–623). Association for Computing Machinery. https://doi.org/10.1145/3442188.3445922

Brown, T. B., Mann, B., Ryder, N., Subbiah, M., Kaplan, J., Dhariwal, P., Neelakantan, A., Shyam, P., Sastry, G., Askell, A., Agarwal, S., Herbert-Voss, A., Krueger, G., Henighan, T., Child, R., Ramesh, A., Ziegler, D. M., Wu, J., Winter, C., … Amodei, D. (2020). Language models are few-shot learners. In H. Larochelle, M. Ranzato, R. Hadsell, M. F. Balcan, & H. Lin (Eds.), *Advances in Neural Information Processing Systems* (Vol. 33, pp. 1877–1901). Curran Associates, Inc. https://proceedings.neurips.cc/paper/2020/hash/1457c0d6bfcb4967418bfb8ac142f64a-Abstract.html

Brunet-Gouet, E., Vidal, N., & Roux, P. (2023). *Can a conversational agent pass theory-of-mind tasks? A case study of ChatGPT with the Hinting, False Beliefs, and Strange Stories paradigms* [Preprint]. HAL. https://hal.science/hal-03991530

Chen, Z., Wu, J., Zhou, J., Wen, B., Bi, G., Jiang, G., Cao, Y., Hu, M., Lai, Y., Xiong, Z., & Huang, M. (2024). ToMBench: Benchmarking theory of mind in large language models. In *Proceedings of the 62nd Annual Meeting of the Association for Computational Linguistics (Volume 1: Long Papers)* (pp. 15959–15983). Association for Computational Linguistics. https://doi.org/10.18653/v1/2024.acl-long.847

Devlin, J., Chang, M.-W., Lee, K., & Toutanova, K. (2019). BERT: Pre-training of deep bidirectional transformers for language understanding. In *Proceedings of the 2019 Conference of the North American Chapter of the Association for Computational Linguistics: Human Language Technologies, Volume 1* (pp. 4171–4186). Association for Computational Linguistics. https://doi.org/10.18653/v1/N19-1423

Elyoseph, Z., Hadar-Shoval, D., Asraf, K., & Lvovsky, M. (2023). ChatGPT outperforms humans in emotional awareness evaluations. *Frontiers in Psychology, 14*, Article 1199058. https://doi.org/10.3389/fpsyg.2023.1199058

Gallese, V., & Goldman, A. (1998). Mirror neurons and the simulation theory of mind-reading. *Trends in Cognitive Sciences, 2*(12), 493–501. https://doi.org/10.1016/S1364-6613(98)01262-5

Gopnik, A., & Wellman, H. M. (1992). Why the child's theory of mind really is a theory. *Mind & Language, 7*(1–2), 145–171. https://doi.org/10.1111/j.1468-0017.1992.tb00202.x

Happé, F. G. E. (1994). An advanced test of theory of mind: Understanding of story characters' thoughts and feelings by able autistic, mentally handicapped, and normal children and adults. *Journal of Autism and Developmental Disorders, 24*(2), 129–154. https://doi.org/10.1007/BF02172093

Heider, F., & Simmel, M. (1944). An experimental study of apparent behavior. *The American Journal of Psychology, 57*(2), 243–259. https://doi.org/10.2307/1416950

Kauf, C., Ivanova, A. A., Rambelli, G., Chersoni, E., She, J. S., Chowdhury, Z., Fedorenko, E., & Lenci, A. (2023). Event knowledge in large language models: The gap between the impossible and the unlikely. *Cognitive Science, 47*(11), Article e13386. https://doi.org/10.1111/cogs.13386

Kosinski, M. (2024). Evaluating large language models in theory of mind tasks. *Proceedings of the National Academy of Sciences, 121*(45), Article e2405460121. https://doi.org/10.1073/pnas.2405460121

Miliani, M., Auriemma, S., Bondielli, A., Chersoni, E., Passaro, L., Sucameli, I., & Lenci, A. (2025). *ExpliCa: Evaluating explicit causal reasoning in large language models* [Preprint]. arXiv. https://doi.org/10.48550/arXiv.2502.15487

Nawaz, S., Lewis, C., Townson, A., & Mei, P. (2023). What does the Strange Stories test measure? Developmental and within-test variation. *Cognitive Development, 65*, Article 101289. https://doi.org/10.1016/j.cogdev.2022.101289

Perner, J., Leekam, S. R., & Wimmer, H. (1987). Three-year-olds' difficulty with false belief: The case for a conceptual deficit. *British Journal of Developmental Psychology, 5*(2), 125–137. https://doi.org/10.1111/j.2044-835X.1987.tb01048.x

Perner, J., & Wimmer, H. (1985). "John thinks that Mary thinks that …": Attribution of second-order beliefs by 5- to 10-year-old children. *Journal of Experimental Child Psychology, 39*(3), 437–471. https://doi.org/10.1016/0022-0965(85)90051-7

Pi, Z., Vadaparty, A., Bergen, B. K., & Jones, C. R. (2024). *Dissecting the Ullman variations with a SCALPEL: Why do LLMs fail at trivial alterations to the false belief task?* [Preprint]. arXiv. https://doi.org/10.48550/arXiv.2406.14737

Posit team. (2024, 2025). *RStudio: Integrated development environment for* R (Version 2025.09.2) [Computer software]. Posit Software, PBC. https://posit.co/products/open-source/rstudio/

R Core Team. (2024, 2025). *R: A language and environment for statistical computing* (Version 4.5.2) [Computer software]. R Foundation for Statistical Computing. https://www.R-project.org/

Radford, A., Narasimhan, K., Salimans, T., & Sutskever, I. (2018). *Improving language understanding by generative pre-training.* OpenAI. https://cdn.openai.com/research-covers/language-unsupervised/language_understanding_paper.pdf

Sabour, S., Liu, S., Zhang, Z., Liu, J. M., Zhou, J., Sunaryo, A. S., Lee, T. M. C., Mihalcea, R., & Huang, M. (2024). EmoBench: Evaluating the emotional intelligence of large language models. In *Proceedings of the 62nd Annual Meeting of the Association for Computational Linguistics (Volume 1: Long Papers)* (pp. 5986–6004). Association for Computational Linguistics. https://doi.org/10.18653/v1/2024.acl-long.326

Sandbrink, K., & Summerfield, C. (2024). Modelling cognitive flexibility with deep neural networks. *Current Opinion in Behavioral Sciences, 57*, Article 101361. https://doi.org/10.1016/j.cobeha.2024.101361

Sap, M., Rashkin, H., Chen, D., Le Bras, R., & Choi, Y. (2019). *SocialIQA: Commonsense reasoning about social interactions* [Preprint]. arXiv. https://doi.org/10.48550/arXiv.1904.09728

Sartori, G., & Orrù, G. (2023). Language models and psychological sciences. *Frontiers in Psychology, 14*, Article 1279317. https://doi.org/10.3389/fpsyg.2023.1279317

Shamay-Tsoory, S. G., Harari, H., Aharon-Peretz, J., & Levkovitz, Y. (2010). The role of the orbitofrontal cortex in affective theory of mind deficits in criminal offenders with psychopathic tendencies. *Cortex, 46*(5), 668–677. https://doi.org/10.1016/j.cortex.2009.04.008

Shapira, N., Levy, M., Alavi, S. H., Zhou, X., Choi, Y., Goldberg, Y., Sap, M., & Shwartz, V. (2024). Clever Hans or neural theory of mind? Stress testing social reasoning in large language models. In *Proceedings of the 18th Conference of the European Chapter of the Association for Computational Linguistics (Volume 1: Long Papers)* (pp. 2257–2273). Association for Computational Linguistics. https://doi.org/10.18653/v1/2024.eacl-long.138

Strachan, J. W. A., Albergo, D., Borghini, G., Pansardi, O., Scaliti, E., Gupta, S., Saxena, K., Rufo, A., Panzeri, S., Manzi, G., Graziano, M. S. A., & Becchio, C. (2024). Testing theory of mind in large language models and humans. *Nature Human Behaviour, 8*(7), 1285–1295. https://doi.org/10.1038/s41562-024-01882-z

Ullman, T. (2023). *Large language models fail on trivial alterations to theory-of-mind tasks* [Preprint]. arXiv. https://doi.org/10.48550/arXiv.2302.08399

van Duijn, M. J., van Dijk, B. M. A., Kouwenhoven, T., de Valk, W., Spruit, M. R., & van der Putten, P. (2023). Theory of mind in large language models: Examining performance of 11 state-of-the-art models vs. children aged 7–10 on advanced tests. In *Proceedings of the 27th Conference on Computational Natural Language Learning* (pp. 389–402). Association for Computational Linguistics. https://doi.org/10.18653/v1/2023.conll-1.25

Vaswani, A., Shazeer, N., Parmar, N., Uszkoreit, J., Jones, L., Gomez, A. N., Kaiser, Ł., & Polosukhin, I. (2017). Attention is all you need. In *Advances in Neural Information Processing Systems* (Vol. 30, pp. 5998–6008). Curran Associates, Inc. https://proceedings.neurips.cc/paper/7181-attention-is-all-you-need

Wang, X., Yuan, P., Feng, S., Li, Y., Pan, B., Wang, H., Hu, Y., & Li, K. (2024). *CogLM: Tracking cognitive development of large language models* [Preprint]. arXiv. https://doi.org/10.48550/arXiv.2408.09150

Wei, J., Wang, X., Schuurmans, D., Bosma, M., Ichter, B., Xia, F., Chi, E. H., Le, Q. V., & Zhou, D. (2022). Chain-of-thought prompting elicits reasoning in large language models. In *Advances in Neural Information Processing Systems*(Vol. 35, pp. 24824–24837). Curran Associates, Inc. https://proceedings.neurips.cc/paper_files/paper/2022/hash/9d5609613524ecf4f15af0f7b31abca4-Abstract-Conference.html

White, S., Hill, E., Happé, F., & Frith, U. (2009). Revisiting the Strange Stories: Revealing mentalizing impairments in autism. *Child Development, 80*(4), 1097–1117. https://doi.org/10.1111/j.1467-8624.2009.01319.x

Xi, Z., Chen, W., Guo, X., He, W., Ding, Y., Hong, B., Zhang, M., Wang, J., Jin, S., Zhou, E., Zheng, R., Fan, X., Wang, X., Xiong, L., Zhou, Y., Wang, W., Jiang, C., Zou, Y., Liu, X., … Gui, T. (2025). The rise and potential of large language model based agents: A survey. *Science China Information Sciences, 68*(2), Article 121101. https://doi.org/10.1007/s11432-024-4222-0

Yuan, Y., Wang, S., Xiong, C., Jiang, Z., & Wu, Y. (2023). *Back to the future: Towards explainable temporal reasoning with large language models* [Preprint]. arXiv. https://doi.org/10.48550/arXiv.2310.01074

Zhao, W. X., Zhou, K., Li, J., Tang, T., Wang, X., Hou, Y., Min, Y., Zhang, B., Zhang, J., Dong, Z., Du, Y., Yang, C., Chen, Y., Chen, Z., Jiang, J., Ren, R., Li, Y., Tang, X., Liu, Z., Liu, P., Nie, J.-Y., & Wen, J.-R. (2023). *A survey of large language models* [Preprint]. arXiv. https://doi.org/10.48550/arXiv.2303.18223

**Appendix A The fictitious animal stories**

Many extremely steep mountains cover an island. Above a certain height no animal can stay conscious because of the lack of oxygen. Floots are the most nimble and shifty species around. They are the only ones capable of climbing above a certain height because of their special hooves. They will, however, often get drowsy during their climbs. Scavengers in the area often feast on floots, even though they are very poor climbers. They can do this without hunting them. QUESTION: How is that possible?

- 2 points—reference to fact that floots get drowsy if they climb high and since the terrain is steep, they may fall. Scavengers may exploit this
- 1 point—reference to drowsiness or the terrain without linking them
- 0 points—any other answer, such as scavengers learning to climb

Pulit are small furry animals that live in large groups. The environment they live in is harsh. It is home to many predators but by staying together as a group, the pulit are able to scare them away. It is a necessary part of their survival. This winter is icy cold, so the pulit keep moving to avoid freezing to death. The ground is very slippery this time of the year. During this time an unusually large number of pulit have lost their life in the hilly area of their country. QUESTION: What happened to them?

- 2 points—reference to the slippery hilly terrain making it difficult for pulit to stay together for protection against predators
- 1 point—reference to slippery terrain without linking it to staying together for protection
- 0 points—any other answer, such as freezing

A very unique type of monkey called wium exist on a small island. Wium have an exceptional ability to regenerate, they can restore their whole limbs if necessary. They have an exceptional instinct to survive, sacrificing anything to do so. Not long ago a new predator entered the ecosystem. It has quickly destroyed the natural balance of the island, making the other small animals that occupied it extinct, leaving their pelts behind. However, the wium have survived alongside the predator, no pelts of theirs are to be found on the ground. After several years the predators and the wium still coexist on the island with almost no wium dying because of them. QUESTION: How is this possible?

- 2 points—reference to the wium sacrificing their limbs to the predators and then regrowing them creating a symbiosis
- 1 point—reference to resilience and ability to regenerate limbs
- 0 points—any other answer, such as learning to evade the predators

The desert is hot and dry, yet it is home to many species; valoos are one of them. They have close familial bonds; the parent animals teach their offspring to hunt and everything that is required for survival, and they would sacrifice their own lives for them. They can differentiate their kin because they share their physical traits. Large types of valoos are fitter for hunting; however, the larger specimens have a much greater appetite. One summer, the desert was especially hot, causing droughts, which led to the animal population almost halving. In the following season, all the prey were caught by the young valoos, most of whom had reached maturity. QUESTION: Why did that happen?

- 2 points – reference to their parents, who needed more food for survival, dying in the draught but by then teaching their children to hunt
- 1 point – the valoos died because they were too big or the young valoos got better at hunting
- 0 points – any other answer such as the young valoos learning to cope by themselves

Riqoue birds are known for their colourful shiny feathers. They are however very rare, because they can't stop flying for longer than a minute. This is due to their breathing system, which needs a lot of fast-moving air, which their lungs can't provide. Recently, however some riqoue birds have escaped the New York city zoo. Surprisingly they have flourished in the bustling city. They can even be seen perching for long durations of time on top of people's helmets. QUESTION: Why can they perch for so long?

- 2 points – helmets are worn by riders of moving vehicles eliminating the need to fly
- 1 point – moving vehicles with no reference to helmets
- 0 points – any other answer such as the rigoue learning to breathe without flying

On a typical day small stika mice play in the open field, basking in the sunlight. At nighttime however they retreat to their underground dwellings, because in the dark predators pose a great threat to them and special wariness is their only protection. They are highly sensitive to changes in light, which is essential for their survival. One October no stika were seen for weeks. There were great storms darkening the sky during this time in the area that destroyed part of the forest. QUESTION: Why were the stika not seen?

- 2 points – The stika hid in their dwellings because the weather brought daytime darkness and the stika were afraid of predators
- 1 point – The stika hid, without mention of the darkness caused by the weather
- 0 points – Any other answer such as the stika dying

**Appendix B Some incorrect answers to the fictitious animal stories**

Here's a possible explanation for what happened to the Pulit: They likely became trapped and exposed in a snowdrift or avalanche. Here's why: Hilly Area: The story mentions a hilly area, which increases the risk of avalanches, especially during heavy snowfall. Winter Conditions: Winter is icy cold, and Pulit need to keep moving to stay warm. If they got caught in a snowdrift or avalanche, they would be unable to move and would quickly freeze. Large Number: The phrase "unusually large number" suggests a significant event, like an avalanche, rather than individual incidents. Let me know if you'd like to explore other possibilities! (Gemma 2)

Based on the story, I would answer that the reason the wium are able to survive alongside the predator without their pelts being found on the ground is because of their exceptional ability to regenerate. The story mentions that the wium have an "exceptional ability to regenerate", which implies that they can restore their whole bodies if necessary. This means that even if a wium is injured or killed by the predator, it will simply regenerate its body and return to normal, without leaving behind a pelt or any other sign of death. In fact, the story suggests that this regeneration ability is so powerful that it allows the wium to survive alongside the predator with almost no casualties. The fact that "no pelts of their were to be found on the ground" implies that the wium are not dying in significant numbers, and therefore do not have any bodies or pelts left behind. Therefore, the answer is: the wium's ability to regenerate allows them to survive alongside the predator without leaving behind any evidence of death. (LLaMa 3.1)

The story describes that during an especially warm summer in the desert, droughts occurred which led to a significant decrease in the population of Valoos - nearly halving it. The following season saw plenty of prey animals; however, all these were caught by the young valoo. This suggests that despite their numbers increasing and reaching maturity after the previous year's drought, they still could not compete with larger types who are fitter for hunting due to their size but have a greater appetite as well. In conclusion, even though the smaller valoo reached maturity in the following season when prey was plentiful, they were unable to effectively hunt and feed because of competition from the larger specimens. The larger Valoos had an advantage due to their size which made them better suited for hunting despite having a bigger appetite as well. (Phi 3)

It is possible that the riqoue birds have adapted to their new environment in New York City. In the city, there may be more air flow and pollution which could provide the birds with the necessary fast-flowing air they need to breathe. Additionally, the birds may have found sources of food that help sustain their energy levels, allowing them to perch for longer durations of time. The urban environment may have provided the riqoue birds with the resources they need to thrive and overcome their natural limitations. (ChatGPT 3.5 turbo)

Based on the story, I would answer: The scavengers are able to eat the floots without them being conscious because the floots become drowsy due to the lack of oxygen at high altitudes, but they do not lose consciousness entirely. My reasoning is as follows: According to the story, animals cannot stay conscious above a certain height on the island due to the lack of oxygen. However, it specifically mentions that the floots will "often get drowsy" during their climbs, implying that while they may be sleepy or disoriented, they are not necessarily unconscious. Therefore, it is possible for scavengers to eat the floots without them being aware of their surroundings or able to defend themselves, but the story does not suggest that the floots are completely out cold. (LLaMa 3.1)

[the stika mice] probably hid from the storms. not safe out when the forest’s getting wrecked. (human)